\documentclass{article}

\PassOptionsToPackage{numbers, compress}{natbib}

\usepackage[preprint]{neurips_data_2024}

\usepackage[utf8]{inputenc} 
\usepackage[T1]{fontenc}    
\usepackage{hyperref}       
\usepackage{url}            
\usepackage{booktabs}       
\usepackage{amsfonts}       
\usepackage{nicefrac}       
\usepackage{microtype}      
\usepackage{xcolor}         
\usepackage{graphicx}
\usepackage{tablefootnote} 
\usepackage{threeparttable}
\usepackage{tipa}
\usepackage{tcolorbox}
\usepackage{multirow}
\usepackage{subcaption}
\usepackage{listings}
\usepackage{array}
\usepackage{makecell}

\newcommand{\depth}[1]{depth$=${#1}}

\newcolumntype{P}[1]{>{\centering\arraybackslash}p{#1}}
\tcbuselibrary{breakable}

\definecolor{b1}{HTML}{1280B0}
\definecolor{b2}{HTML}{25537D}
\colorlet{punct}{red!60!black}
\definecolor{background}{HTML}{EEEEEE}
\definecolor{delim}{RGB}{20,105,176}
\definecolor{codegreen}{rgb}{0,0.6,0}
\definecolor{codegray}{rgb}{0.5,0.5,0.5}
\colorlet{numb}{magenta!60!black}
\lstdefinelanguage{json}{
    basicstyle=\normalfont\ttfamily,
    numbers=left,
    numberstyle=\scriptsize,
    stepnumber=1,
    numbersep=8pt,
    showstringspaces=false,
    breaklines=true,
    frame=lines,
    backgroundcolor=\color{background},
    commentstyle=\color{codegreen},
    literate=
     *{0}{{{\color{numb}0}}}{1}
      {1}{{{\color{numb}1}}}{1}
      {2}{{{\color{numb}2}}}{1}
      {3}{{{\color{numb}3}}}{1}
      {4}{{{\color{numb}4}}}{1}
      {5}{{{\color{numb}5}}}{1}
      {6}{{{\color{numb}6}}}{1}
      {7}{{{\color{numb}7}}}{1}
      {8}{{{\color{numb}8}}}{1}
      {9}{{{\color{numb}9}}}{1}
      {:}{{{\color{punct}{:}}}}{1}
      {,}{{{\color{punct}{,}}}}{1}
      {\{}{{{\color{delim}{\{}}}}{1}
      {\}}{{{\color{delim}{\}}}}}{1}
      {[}{{{\color{delim}{[}}}}{1}
      {]}{{{\color{delim}{]}}}}{1},
}

\definecolor{isarblue}{HTML}{006699}
\definecolor{isarfaintblue}{rgb}{0.0, 0.75, 1.0}
\definecolor{isargreen}{HTML}{009966}
\definecolor{red}{HTML}{990000}
\definecolor{patriarch}{rgb}{0.5, 0.0, 0.5}
\lstdefinelanguage{isabelle}{%
    backgroundcolor=\color{background},
    keywords=[1]{type_synonym,datatype,fun,abbreviation,definition,proof,lemma,theorem,qed,corollary,have,hence,also,finally,ultimately,moreover,using,\{},
    keywordstyle=[1]\bfseries\color{isarblue},
    keywords=[2]{where,assumes,shows,fixes,and},
    keywordstyle=[2]\bfseries\color{isargreen},
    keywords=[3]{if,then,else,case,SOME,let,in,O},
    keywordstyle=[3]\color{isarblue},
    keywords=[4]{ATP},
    keywordstyle=[4]\it\color{patriarch},
    keywords=[5]{show,assume,obtain},
    keywordstyle=[5]\bfseries\color{isarfaintblue},
}

\lstdefinestyle{isabelle}{%
  language=isabelle,
  escapeinside={\&}{&},
  columns=fixed,
  extendedchars,
  basewidth={0.5em,0.45em},
  basicstyle=\singlespacing\ttfamily\small,
  mathescape,
  morecomment=[s][\bfseries\color{red}]{(*}{*)},
  morecomment=[l][\bfseries]{####},
}

\lstdefinestyle{mystyle}{
    basicstyle=\normalfont\fontfamily{ptm},
    breaklines=true,
}

\usepackage{soul}
\colorlet{lightdelectricgray}{codegray!20}
\newcommand{\hldbGray}[1]{%
    {%
    \sethlcolor{lightdelectricgray}%
    \hl{#1}%
    }%
}

\hypersetup{
    colorlinks=true,
    linkcolor=black,
    filecolor=black,      
    urlcolor=magenta,
    citecolor=b1,
}

\newcommand{\ourmethod}[0]{\textsc{FVEL}}
\newcommand{\ourdataset}[0]{\textsc{FVELer}}

\title{
\ourmethod: Interactive Formal Verification Environment with Large Language Models via Theorem Proving
}

\author{
Xiaohan Lin\textsuperscript{1}\thanks{\quad Equal contribution.} ~~
Qingxing Cao\textsuperscript{1}\footnotemark[1] ~~
Yinya Huang\textsuperscript{2}\footnotemark[1] ~~
Haiming Wang\textsuperscript{3}\footnotemark[1] ~~
Jianqiao Lu\textsuperscript{4} \\
{\bf Zhengying Liu\textsuperscript{5}} ~~
{\bf Linqi Song\textsuperscript{2}} ~~
{\bf Xiaodan Liang\textsuperscript{1,6,7}\thanks{\quad Corresponding author.}}\\
$^1$Shenzhen Campus of Sun Yat-sen University ~~
$^2$City Univeristy of Hong Kong \\
$^3$Sun Yat-sen University ~~
$^4$The University of Hong Kong \\
$^5$Huawei Noah's Ark Lab ~~
$^6$MBZUAI ~~
$^7$DarkMatter AI Research 
}

\begin{document}

\maketitle

\begin{abstract}
Formal verification (FV) has witnessed growing significance with current emerging program synthesis by the evolving large language models (LLMs). 
However, current formal verification mainly resorts to symbolic verifiers or hand-craft rules, resulting in limitations for extensive and flexible verification. 
On the other hand, formal languages for automated theorem proving, such as Isabelle, as another line of rigorous verification, are maintained with comprehensive rules and theorems.  
In this paper, we propose
\ourmethod\footnote{
\ourmethod: Pronounced as fuel. 
\ourdataset: Pronounced as fueler. \ourmethod\ \textbf{\underline{e}}nvironment \textbf{\underline{r}}esource.
},
an interactive \textbf{\underline{F}}ormal \textbf{\underline{V}}erification \textbf{\underline{E}}nvironment with \textbf{\underline{L}}LMs. 
Specifically, \ourmethod\ transforms a given code to be verified into Isabelle, and then conducts verification via neural automated theorem proving with an LLM. 
The joined paradigm leverages the rigorous yet abundant formulated and organized rules in Isabelle and is also convenient for introducing and adjusting cutting-edge LLMs. 
To achieve this goal, we extract a large-scale \ourdataset\footnotemark[3].
The \ourdataset\ dataset includes code dependencies and verification processes that are formulated in Isabelle, containing 758 theories, 29,125 lemmas, and 200,646 proof steps in total with in-depth dependencies.
%
We benchmark \ourdataset\ in the \ourmethod\ environment by first fine-tuning LLMs with \ourdataset\ and then evaluating them on Code2Inv and SV-COMP.
The results show that \ourmethod\ with \ourdataset\ fine-tuned Llama3-8B solves 17.39\% (69$\to$81) more problems, and Mistral-7B 12\% (75$\to$84) more problems in SV-COMP. And the proportion of proof errors is reduced.
%
Project page: \url{https://fveler.github.io/}.
\end{abstract}

\section{Introduction}
%
Formal verification (FV), or automated program verification \cite{DBLP:conf/cav/SiNDNS20,DBLP:conf/tacas/Beyer23} checks if a code meets a specific demand and is correct to implement. 
As the code synthesis ability of current models \cite{DBLP:journals/corr/abs-2303-08774,DBLP:journals/corr/abs-2306-08568,DBLP:journals/corr/abs-2401-14196} evolves rapidly, 
there is a growing demand for automated verification of diverse and abundant synthesis programs. 
However, current formal verification mainly resorts to symbolic verifiers \cite{DBLP:conf/tacas/HeizmannCDEHLNSP13,DBLP:conf/kbse/GadelhaMMC0N18,DBLP:conf/tacas/KroeningT14} or hand-craft rules \cite{wu2024lemur}. 
However, symbolic verification can not leverage the advanced reasoning ability of current large language models (LLMs), while hand-craft rules with limited execution on specific code cases have restricted abilities to general verification.

On the other hand, automated theorem proving (ATP) \cite{DBLP:conf/iclr/ZhengHP22,DBLP:conf/cpp/X20,DiskPaxos-AFP} is a line of work on rigorous verification with formal languages (e.g., Isabelle \cite{DBLP:books/sp/Paulson94}, Lean \cite{DBLP:conf/cade/MouraKADR15}) and interactive proof environments (e.g., PISA \cite{jiang2021lisa}, LeanDojo \cite{DBLP:conf/nips/YangSGCSYGPA23}). 
Such formal languages and toolkits maintain corresponding libraries with a large number of human-written and checked theorems and rules, which are provided as pre-training materials for many large language models \cite{DBLP:journals/corr/abs-2303-08774,DBLP:journals/corr/abs-2307-09288,DBLP:journals/corr/abs-2310-06825}.
The ATP formulation and rules have strong expressiveness and, therefore have a great potential for describing formal verification problems and requests. As a result, the verification can be implemented under a rigorous, step-wise, and interactive ATP environment. 
Moreover, the pre-trained formal reasoning capabilities within LLMs and their potential to solve formal verification problems are underexplored. 

\begin{figure}[!t]
    \centering
    \includegraphics[width=1\textwidth]{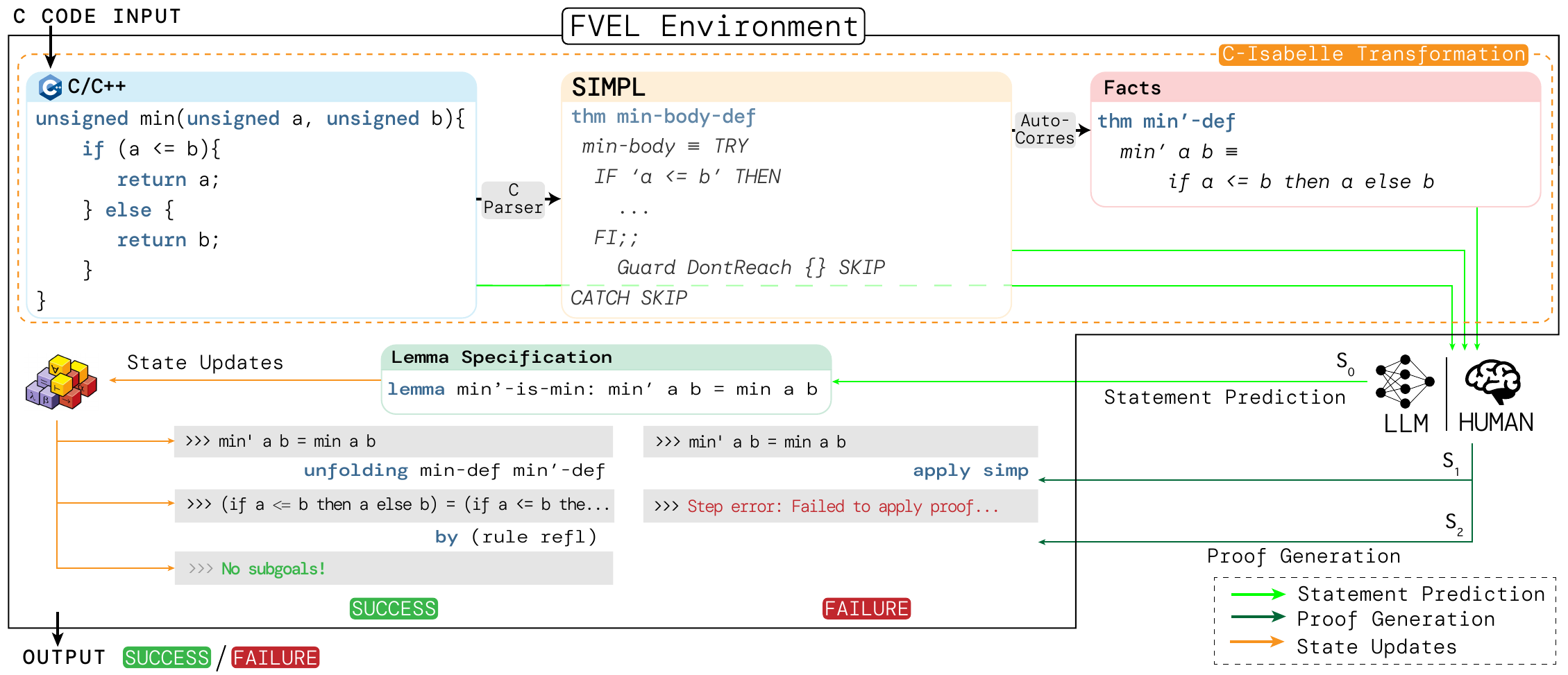}
    \caption{
        \ourmethod\ workflow.
        \ourmethod\ takes a C code as input, parses it into Isabelle definition, 
        and then conducts interactive formal proving with
        \ourmethod-LLM/human via outputting proof state and receiving generated proof.
    }
    \label{fig:framework}
\end{figure}

To take one step towards this goal, 
this paper proposes \ourmethod, a new formal verification environment interacting with LLMs via automated theorem proving processes. 
Figure~\ref{fig:framework} demonstrates an overview of \ourmethod. 
Specifically, the \ourmethod\ environment takes as input a code to be verified, converts the code into Isabelle formulation, and generates a lemma in Isabelle followed by a whole proof to the lemma. \ourmethod\ then outputs the proof result (succeed or failed being proved) as an indication of the code verification result. 
\ourmethod\ interacts with an LLM by initially providing the converted Isabelle formulation to the LLM and then receiving the derived lemma on the code specification.
The interaction is then continued by the LLM generating proof states and the \ourmethod\ environment providing feedback via prover information in the PISA environment \cite{jiang2021lisa}, such as cheating keywords \texttt{sorry} or \texttt{opps} and other error messages. 
As a result, a user provides her code to be verified to \ourmethod, and then she will receive the verification result and intermediate proving information. 
Note that we follow previous works \cite{DBLP:conf/kbse/GadelhaMMC0N18,wu2024lemur} to investigate \ourmethod\ on C code verification in this paper. We remain the extension of \ourmethod\ to support more program languages as a near future work. 

To implement the \ourmethod\ environment, we extract and cleanse a large-scale \ourdataset\ dataset with deep dependencies, which can be applied as both a fine-tuning resource and evaluation benchmark. 
The \ourdataset\ dataset has two main components: C code dependencies formulated by Isabelle theories, and Isabelle lemmas with their step-wise proof states. 
\ourdataset\ then includes 758 theories with 29,125 lemmas and 200,646 proof steps. 
The dataset is then randomly split according to lemmas, resulting in training/validation/test/test-hard sets. The test-hard set data have dependencies that are challenging to find. 
Statistical analysis shows that \ourdataset\ data comprehensively covers diverse dependency depths and has a remarkable number of data with very deep dependencies. For example, over 50\% of lemmas have a depth greater than 78, while the deepest dependency is 156.

We then benchmark \ourdataset\ in the \ourmethod\ environment 
on the Code2Inv \cite{DBLP:conf/cav/SiNDNS20} and SV-COMP \cite{DBLP:conf/tacas/Beyer23} benchmarks.
After fine-tuning on \ourdataset, 
Mistral-7B~\cite{DBLP:journals/corr/abs-2310-06825} and Llama3-8B\footnote{\url{https://github.com/meta-llama/llama3}} are observed performance improvements on both benchmarks. 
For example, Llama3-8B solves 81 out of 1,000 SV-COMP problems, achieving a 17.39\% improvement,
and Mistral-7B improves by 12\%.
Moreover, 
ablation study on statement and proof errors during \ourmethod\ verification shows that after fine-tuning with \ourdataset, the proportion of proof errors is reduced, indicating the benefits of \ourmethod\ and \ourdataset.
%
The contributions of this paper are summarized as follows:
\begin{enumerate}
    \item We introduce \ourmethod, an interactive formal verification environment with LLMs that leverages neural ATP advances including formulation, theorems, models, and prover.     
    \item We extract and cleanse a large-scale \ourdataset\ with 758 theories, 29,125 lemmas, and 200,646 proof steps in total that contain deep dependencies. We split \ourdataset\ into training/validation/test/test-hard sets as fine-tuning resources and an evaluation benchmark.    
    \item We apply \ourmethod\ with several \ourdataset\ fine-tuned LLMs. The results show that \ourmethod\ with \ourdataset\ fine-tuned LLMs show performance improvements on representative code verification benchmarks, and the proof errors are reduced. The results indicate the benefits of \ourmethod\ and \ourdataset.    
\end{enumerate}

\section{Related Works}
\noindent\textbf{Formal Verification.}
Formal verification (FV), or automated program verification \cite{DBLP:conf/cav/SiNDNS20,DBLP:conf/tacas/Beyer23}, is the task of verifying if a given code fulfills specific requirements. 
One line of work \cite{DBLP:conf/tacas/HeizmannCDEHLNSP13,DBLP:conf/kbse/GadelhaMMC0N18,DBLP:conf/tacas/KroeningT14} resort to reducing the code into candidate loop invariant and then using satisfiability modulo theories (SMT) solver for post-hoc verification. 
Different methods are proposed to improve the loop invariant inference, including decision tree \cite{DBLP:journals/corr/KrishnaPW15}, reinforcement learning \cite{DBLP:conf/issta/Yu0023,DBLP:conf/nips/SiDRNS18}, and neural network \cite{DBLP:conf/iclr/RyanWYGJ20}. 
However, finding or generating accurate loop invariants remains challenging, which hinders the preciseness of the verification. 
Moreover, symbolic SMT solvers are time-consuming and uneconomical when there is a large amount of code to be verified. 
The other line of work tries to introduce LLMs to solving formal verification. 
For using LLMs to find loop invariants, Loopy \cite{DBLP:journals/corr/abs-2311-07948} prompts LLMs to exhaustively generate candidate invariants and include a repair procedure to improve the variants by an SMT solver. 
For using LLMs to perform the program verification, 
Lemur \cite{wu2024lemur} proposes to integrate LLMs in formal verification by transforming the program invariants into deductively verified sub-goals, appearing to be most relevant to our work. However, they hand-craft a proof system with solely 8 rules without a demonstration of its completeness. Therefore, the expressiveness of this hand-craft system is unclear. 
In this paper, we propose a new formal verification environment that interacts with large language models to leverage their theorem-proving ability and also the rigorous validation by automated theorem provers. The environment thus leverages the corresponding extensive rule and theorem libraries. 

\noindent\textbf{Automated Theorem Proving with LLMs.}
The field of automated theorem proving (ATP) \cite{schulz2002brainiac,DBLP:conf/cav/KovacsV13,DBLP:journals/jar/ChouGZ00,DBLP:journals/jsc/OttenB03,DBLP:journals/corr/abs-2404-09939} has developed formal languages such as first-order logic (FOL) and higher-order logic (HOL) to describe mathematical problems, theorems, and solution processes, allowing deductive reasoning to achieve the final answer or proof with rigorous stepwise validation.
Interactive theorem proving (ITP) then introduces interactive proof assistances \cite{DBLP:books/sp/Paulson94,DBLP:conf/cade/MouraKADR15,coq1996coq,megill2019metamath} and automates the validation process with machine learning methods \cite{DBLP:journals/corr/abs-2009-03393,DBLP:conf/iclr/HanRWAP22,DBLP:conf/nips/JiangLTCOMWJ22,DBLP:conf/acl/WangYLSYXXSLL0L23}. 
Furthermore, recent studies explore the integration of large language models and theorem proving \cite{jiang2021lisa,DBLP:conf/nips/YangSGCSYGPA23,wang2024legoprover,huang2024mustard,lu2024processdriven}. 
For example, PISA \cite{jiang2021lisa} introduces an environment that allows language models to interact with an Isabelle server, which are able to mine 183k lemmas and theorems from the Isabelle libraries.
LeanDojo \cite{DBLP:conf/nips/YangSGCSYGPA23}, on the other hand, is a Lean environment that enables interaction between the language models and the Lean prover with fine-grained annotations of premises in proofs and an LLM-based theorem prover. 
Such interactive proving systems leverage both the abundant libraries of theorems and rules and advanced performances of LLMs, which is promising for formalized applications such as formal verification.
To this end, this paper investigates a novel LLM interactive environment that advances formal verification. The environment thus also helps solve automated theorem proving tasks.

\section{\ourmethod: Interactive FV Environment with LLMs}
%
%
%
\noindent\textbf{Workflow of \ourmethod.}
Figure~\ref{fig:framework} demonstrates \ourmethod. 
The main idea of \ourmethod\ is to provide an interactive environment with large language models (LLMs) that leverage rigorous theorem-proving processes. 
The input of \ourmethod\ environment is a code to be verified. 
Specifically, we follow previous studies \cite{DBLP:conf/tacas/HeizmannCDEHLNSP13,wu2024lemur} to verify C code and conduct a pilot study on our new framework.
Moreover, the input format is flexible as one can choose to input an ensemble of C code and its corresponding SIMPL and/or Isabelle content as supplements. 
The output of \ourmethod\ is the code verification result, i.e., success or failure.

\ourmethod\ interacts with a large language model to achieve the verification. 
At the initial step of interaction ($S_0$ in Figure~\ref{fig:framework}), \ourmethod\ transforms the input C code into facts, and then provides the facts to the LLM. 
The LLM then generates a lemma in Isabelle \cite{DBLP:books/sp/Paulson94} as a formal description of the code specification. 
In this step, a code verification problem is transformed into an ATP problem. As a result, \ourmethod\ can leverage the LLMs theorem-proving techniques and rigorous ATP validation. 
At the follow-up interaction steps ($S_i, i\ge 1$ in Figure~\ref{fig:framework}), the LLM is prompted to generate proof steps, while \ourmethod\ incorporates an Isabell prover to provide feedback such as error messages to the LLM. 
The process terminates until a whole proof is generated.
If the proof success in proving the lemma, \ourmethod\ outputs ``success'', otherwise outputs ``failure''.

\noindent\textbf{Applying \ourmethod.}
The current version of \ourmethod\ supports code verification in C language. 
We leave the generalization of \ourmethod\ to other program languages as a near future work. 
To apply \ourmethod, a user prepares her C code and passes it to \ourmethod. 
The user can customize her LLM for \ourmethod. 
Therefore, \ourmethod\ adjusts to cutting-edge LLMs with strong theorem-proving ability and customized LLMs. 
The user then gets the ``success'' or ``failure'' feedback regarding the verification result from \ourmethod. 
Furthermore, the intermediate proof states and prover messages provide further information about the verification. 

\noindent\textbf{Environment Implementation.}
We perform the c code transformation with the C-Parser~\cite{norrish2016cparser} and AutoCorres~\cite{greenaway2016autocorres} and construct the environment based on Isabelle-scala\footnote{\url{https://github.com/dominique-unruh/scala-isabelle}} and PISA~\cite{jiang2021lisa}.
C-Parser can translate a large subset of C99 code into the imperative language SIMPL. For every function in the C source file, it generates a corresponding Isabelle definition literally without omitting details of the C language. AutoCorres can further simplify and abstract the generated SIMPL language, producing a higher-level functional specification that is easier to reason by humans. We provide the simplified Isabelle definition to LLMs to better align with human interactive proving with Isabelle.
%
%
Specifically, Given the c source file, we use the PISA to set up the Isabelle process by including the directories of C-parser and AutoCorres in the Isabelle ``sessionRoots'', and setting the ``\texttt{workingDirectory}'' to the C file. Then we initial the Isabelle state by importing the C-parser and AutoCorres tools. Lastly, we use PISA to interact with the Isabelle process, invoke tools to translate the C code, and then extract the fact definition ``{c file name}.{function name}'\_def'' after unfolding it in Isabelle.
The extracted definition can be passed to LLMs, and LLMs can generate lemma specifications and interact with Isabelle prover in this setup process.

\begin{figure}[!t]
    \centering
    \includegraphics[width=1\textwidth]{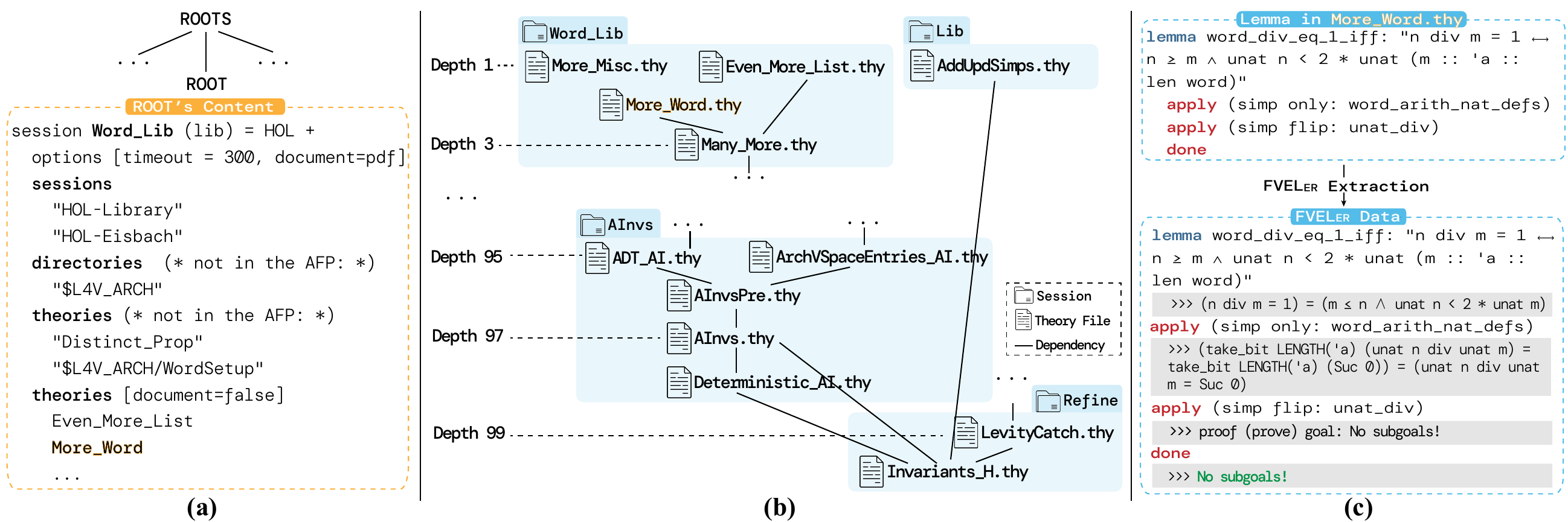}
    \caption{
        (a) SeL4 ROOT file structure. It provides an example ROOT file content for the session \texttt{Word\_Lib}.
        (b) Theory dependency graph. Each theory file is grouped by the Session.
        (c) Step-wise lemmas extraction.
    }
    \label{fig:dataset}
\end{figure}

\section{\ourdataset: Benchmarking \ourmethod}
\subsection{\ourdataset\ Overview}
\ourdataset\ contains transformed Isabelle theories and lemmas from C codes that support the \ourmethod\ environment for C code verification.
\ourdataset\ has two main components:
(1) Theories dependencies. A resource for dependencies among theories, lemmas, and c code specified by SeL4 verification. 
These data provide the ground-truth seL4 premises for proving the current lemma and enable a model to retrieve related statements or proof context at both the training and testing stages.
(2) Lemmas from theories with their Isabelle proof states. The step-wise lemmas with multiple proof states that support the Isabelle proving process in \ourmethod. 
These data on the one hand enhance LLMs with search-based/step-wise ATP while interacting with \ourmethod, and on the other hand, provide a benchmark for interactive formal verification. 
Figure~\ref{fig:dataset} illustrates the construction processes of each component.

In the following, we first introduce the preliminary for \ourdataset\ construction (Section~\ref{sec:prep}), 
then introduce the construction of the two components in \ourdataset: 
(1) the extraction of C-Code Dependencies by Isabelle Theories (Section \ref{sec:theory_dependency_graph}) and
(2) the extraction of step-wise lemmas (Section \ref{sec:step_state_extraction}). 
We then demonstrate \ourdataset\ statistics and distribution in Section \ref{sec:data_stat}.

\subsection{Preparation}
\noindent\textbf{Data Source.}
\label{sec:prep}
SeL4\footnote{The l4v library which contains the proofs for the SeL4 kernel are licensed under GPL version 2.}~\cite{DBLP:conf/sosp/KleinEHACDEEKNSTW09} is a system microkernel with comprehensive formal verification. 
Its implementation verification against safety and security specifications contains multi-level formal proof manually written in Isabelle, including abstract specification and concept level to concrete implementation level.
Since the open-source seL4 verification contains high-quality and multi-level proof following human reasoning, we choose seL4 as \ourdataset\ data source. 
Figure~\ref{fig:dataset}(a) demonstrates the relations amount session, ROOT files, and lemmas in seL4.

\noindent\textbf{SeL4 Session.}
In seL4, an Isabelle session contains a group of theory files that focus on proofing one concept or topic, similar to the package in a programming language. Since the formal verification of seL4 is a large project that involves various aspects, 
different sessions are used
to define code specifications, construct intermediate definitions, and process C code semantics. 
Isabelle can build a session into a binary file called ``heap image'' that can be fast-loaded for processing other theories. 

\noindent\textbf{ROOT Files.} The ROOT files contain all the listed ROOTs that should be built by Isabelle. ROOT files instruct Isabelle on how to build the sessions and verify the theories. Each session in a ROOT file contains its names, parent sessions, entry theories, and directories of theory files. We use such information to recursively construct the dependency graphs and set up the Isabelle environment to extract step-wise proof states. 

\noindent\textbf{Theory and Lemma.} A Theory file contains the necessary context and concrete proof for Isabelle to formally verify the target lemmas. The context includes importing other theories, defining intermediate symbols, and giving concrete lemma statements and proof. A lemma is a statement that relates to the functionality demands of the codes. 
In \ourmethod, the goal of formal verification is to generate the correct proof of these lemmas.

\subsection{\ourdataset\ Construction: C-Code Dependencies by Isabelle Theories}
\label{sec:theory_dependency_graph}
The dependencies are all formulated and saved in Isabelle.
The extraction of the dependencies is via constructing a theory dependency graph.
Figure~\ref{fig:dataset}(b) illustrates the theory dependency graph. 
This graph nodes are the \texttt{.thy} theory files in seL4 while the edges are the \texttt{import} relationships between the theory files.
It traces multi-hop dependency relationships by \texttt{import} among the Isabelle lemmas within the theory files. 
With the theory dependency graph, it is convenient to locate and extract multi-depth lemmas and their corresponding proofs. 

While constructing the dependency graph,
we first traverse all ROOT files according to the file order specified in seL4,
and then parse the session and corresponding theories recursively to obtain the dependency relationship.
%
Specifically,
the graph construction is started by sequentially parsing the ROOT files in the seL4 ROOTS file. For each session, we match the keywords to extract its name, its parents, and its directories. After extracting all session information, we traverse the ROOTS again and parse the theory files under the ``theories'' keywords. We parse the string between ``imports'' and ``begin'' keywords to extract the dependency relationships of these parent theories and parse these theories recursively to form a graph of other theories given current or other session information.
After traversing all sessions, we construct a dependency graph among sessions and theories, which can be used to provide dependent proof context or premise when generating formal verification.

\begin{table}[t]
  \caption{
      \ourdataset\ Statistics.
      A \textit{theory} is a \texttt{.thy} file in seL4 that contains multiple \textit{lemmas}. 
      Each \textit{lemma} has multiple \textit{proof steps}.
      The train/val/test/test-hard data split is based on \textit{lemmas}. 
  }
  \label{tab:stat}
  \centering
  \vspace{3pt}
  \begin{threeparttable}  
  \begin{tabular}{lrrrrr}
    \toprule
     & Total & Train & Val & Test & Test-Hard \\
    \midrule
    $\triangleright$ \textit{Theory} \\
    \midrule
    Number of Theories & 758 & - & - & - & - \\
    Average depth$^{*}$ & - & 73.687 & 73.732 & 73.958 & 31.476 \\
    Maximum depth & 156 & 156 & 156 & 156 & 115 \\
    \midrule
    $\triangleright$ \textit{Lemma} \\
    \midrule
    Number of Lemmas & 29,125 & 26,081 & 1,115 & 1,077 & 852 \\
    \midrule
    $\triangleright$ \textit{Proof Step} \\
    \midrule
    Number of proof steps$^{**}$ & 200,646 & 179,289 & 8,035 & 8,678 & 4,644 \\
    Average proof steps & - & 6.874 & 7.206 & 8.057 & 5.450 \\
    Maximum proof steps & 963 & 944 & 404 & 963 & 107 \\
    \bottomrule
  \end{tabular}
  \begin{tablenotes} 
      \footnotesize
      \item[*] Depth: Degree of the theory dependency graph by \texttt{import} relationship. 
      \item[**] Proof step: A single step in Isabelle producing a valid statement for interaction."
  \end{tablenotes}
  \end{threeparttable}
\end{table}

\subsection{\ourdataset\ Construction: Step-Wise Lemmas}
\label{sec:step_state_extraction}
For extracting the lemmas and also saving their dependencies by theory files and their proof states, we leverage the PISA \cite{jiang2021lisa} environment. 
We initial the PISA environment and parse all theory files based on the session information the theory dependency graph developed in Section~\ref{sec:theory_dependency_graph}. 
Specifically,
As shown in Figure~\ref{fig:dataset}(c),
we first build the seL4 formal verification project\footnote{\url{https://github.com/seL4/l4v}} and obtain the sessions' binary heap images. Then given each theory file, we modify the PISA environment to include and load all dependent sessions, setting the working directories to the processed theory files, and then temporarily copying the files from session directories to the current one, such that the Isabelle process can correctly import all dependent theory.
Lastly, we use PISA to parse the theory file into multi-step and perform step-wise interaction with Isabelle. For each step, Isabelle will return a proof state and we store the step and proof state as a step-wise training sample.
We traverse the seL4 verification projects and extract most of the theory files. Specifically, we omit some experimental theory files that can not be verified by Isabelle or failed when interacting with PISA. 
We also omit the sessions for documentation, C parser \cite{norrish2016cparser} and AutoCorres \cite{greenaway2016autocorres} as they do not contain lemmas that are relevant to formal proving.

\begin{figure}[t]
    \centering
    \begin{subfigure}[]{\textwidth}
        \includegraphics[width=1\textwidth]{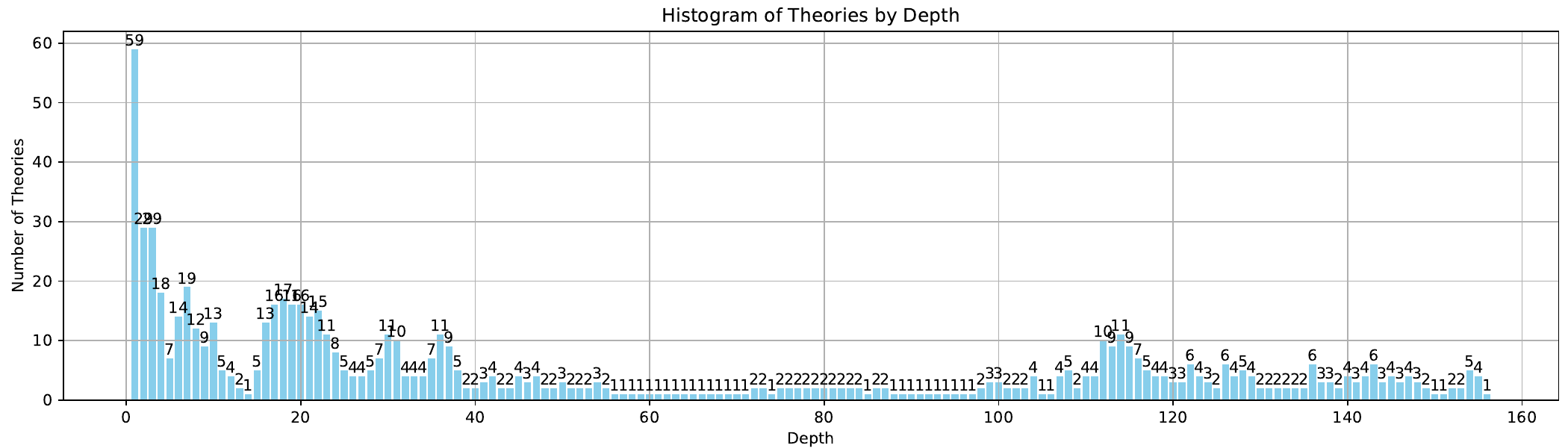}
        \caption{\small Distribution of dependency by theory.}   
        \label{fig:stat_theorem_depth}
    \end{subfigure}
    \begin{subfigure}[]{\textwidth}
        \includegraphics[width=1\textwidth]{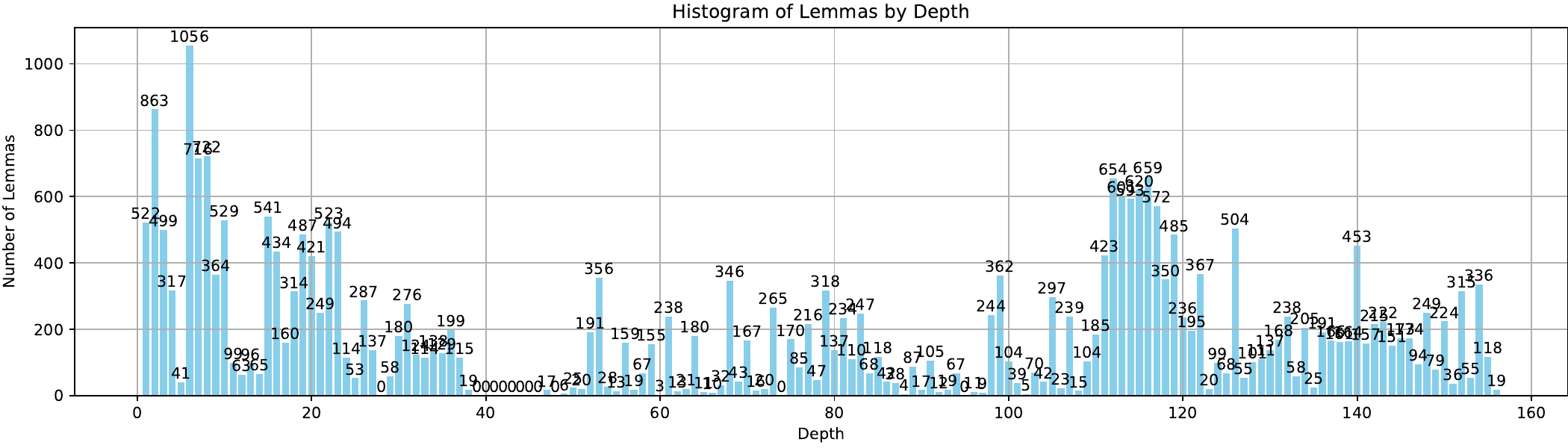}
        \caption{Distribution of dependency by lemma.} 
        \label{fig:stat_lemma_depth}
    \end{subfigure}
    \caption{
        The \ourdataset\ dependency distributions by theory and lemma, respectively.
    }
\end{figure}

\subsection{\ourdataset\ Splits, Statistics, and Distributions}  
\label{sec:data_stat}

\noindent\textbf{Splits.}
We randomly split \ourdataset\ according to lemmas, resulting in a training set, a validation set, a test set, and an especially selected test-hard set. 
The test-hard set is selected from those lemmas in the three sessions ``SysInit'', ``SysInitExamples'', ``LibTest''. Such lemmas are in higher depths in the dependency relationship, therefore they have less \texttt{import} relationships by other theories. 

\noindent\textbf{Statistics.}
Table~\ref{tab:stat} demonstrates the number of samples in \ourdataset\ and each data split. 
\ourdataset\ in total contains 758 Isabelle theories, with 29,125 lemmas and 200,646 proof steps. 
The average dependency depths among the theories range from 31 to 73. 
The maximum dependency path reaches a depth of 156. 
The average proof step ranges from 5 to 8, while the maximum of proof steps in a lemma reaches 963.
In general, \ourdataset\ is a large-scale dataset with deep dependencies among the Isabelle theorems and lemmas that fit C code formulation. It thus supports the interactive C code verification with a theorem-proving LLM. 

\noindent\textbf{Distribution of Dependency by Theory.}
We quantify the dependency by ``depth'', which is the degree of the theory dependency graph by the \texttt{import} dependency relationship among the theory files, as introduced in Section~\ref{sec:theory_dependency_graph}.
Figure~\ref{fig:stat_theorem_depth} demonstrates the distribution of theories by the depth of dependency relationship. 
Besides the number of theories in \depth{1} is the highest 59 followed by \depth{2} and \depth{3} with 29 theories, respectively, small peaks are observed in multiple depth levels. 
For example, \depth{7} has 19 theories, \depth{16} to 22 have around 15 theories, and there are still 11 theories that have \depth{36}. 
Most impressively,
\depth{112} to 115 appear to have on average around 10 theories. 
As a result, \ourdataset\ has very in-depth and comprehensive dependencies information, 
which can be beneficial for not only code verification with dependencies but also multi-step ATP. 

\noindent\textbf{Distribution of Dependency by Lemma.}
Figure~\ref{fig:stat_lemma_depth} demonstrates the distribution of 29,125 lemmas by depth.
That is, each lemma belongs to one of the 758 theories whose depth in its dependency is calculated here. Therefore, in Figure~\ref{fig:stat_lemma_depth} we observe a more fine-grained dependency distribution within the theory files. 
It is shown that lemmas with deep dependency are widely distributed.
Lemmas with depth$\ge$78 are 14,668, over 50\% of all lemmas. 
For example, \depth{116} there are 659 lemmas. 
Moreover, there are also 11,518 lemmas with shorter \depth{1} to 40.
%
Besides, a curious observation is that \depth{39} to 46 are not found in lemmas. 
Therefore, \ourdataset\ widely supports verification with diverse depths of dependency. 

\begin{figure}[t]
    \centering
    \includegraphics[width=0.97\textwidth]{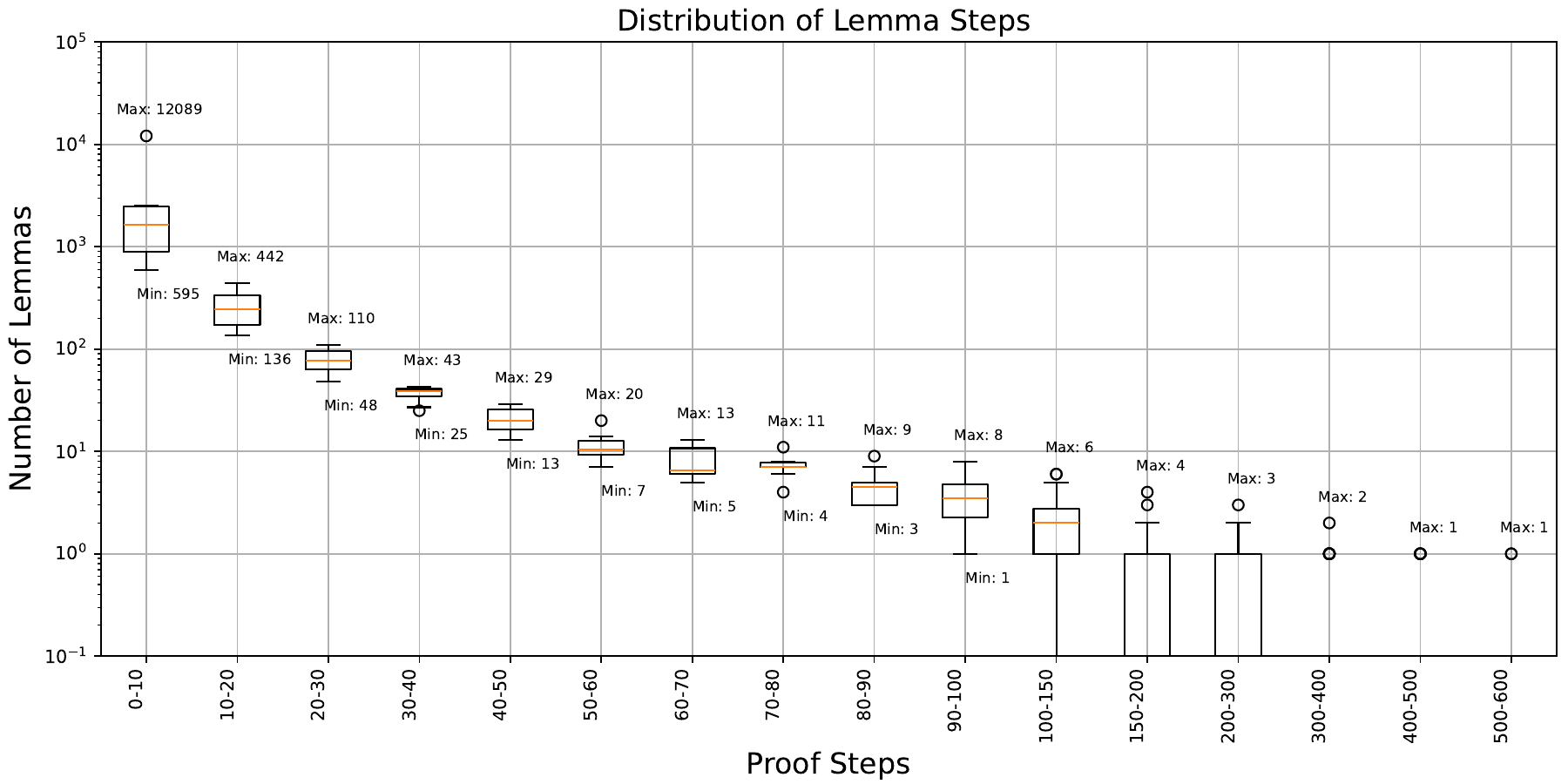}
    \caption{
        The \ourdataset\ lemma distribution over step intervals. 
        We adjust the range by setting the y-axis to a logarithmic scale.
    } 
    \label{fig:stat_lemma_steps}
\end{figure}

\noindent\textbf{Distribution of Lemma Steps.}
One proof step in a lemma is from a current proof state to the next which produces a sound statement for interaction in PISA. 
Figure~\ref{fig:stat_lemma_steps} demonstrates the distribution of intermediate proof steps of the 29,125 lemmas.
It is indicated that the number of proof steps is dramatically different from that of lemmas. 
12,089 out of the 29,125 lemmas can be proved via one proof step. 
Proof steps between 2 and 10 there are 12,957 lemmas. 
Therefore, over 85\% of the lemmas in \ourdataset\ can be proved within 10 steps. 
Moreover, 28,954 out of the 29,125 lemmas can be proved within 100 steps. 
Therefore \ourdataset\ is more helpful for verification within 100 ATP steps, {which is sufficient for covering most of the cases in practice}.

\section{Benchmark Study}
\subsection{Setup}
\noindent\textbf{Dataset.}
We benchmark \ourdataset\ in the \ourmethod\ environment
on Code2Inv~\cite{DBLP:conf/cav/SiNDNS20} and SV-COMP~\cite{DBLP:conf/tacas/Beyer23}.
The Code2Inv dataset contains 133 programs in c, and the SV-COMP dataset is from the Software-Verification Competition with over 23k c programs.
Since C-parser supports only part of the C99 standard, we normalize the C code to make C-parser work properly.
For more preprocessing and implementation details, please refer to the supplementary materials. 

\noindent\textbf{Fine-tuning.}
We use the training set of \ourdataset\ to fine-tune language models. In this study, we employ LORA~\cite{hu2021lora} to fine-tune two most advanced open-source large language models which excel in mathematical reasoning and code generation: Llama-3-8B-instruct\footnotemark[4] and Mistral-7B-Instruct-v0.2 \cite{DBLP:journals/corr/abs-2310-06825}. We convert the training data into the alpaca format, where all training samples use the same instruction, the input is the lemma specification, and the output is a complete proof written in Isabelle. 

\noindent\textbf{Inference.}
During inference, we transfer the input c-code functions into Isabelle facts in \ourmethod\ environment, requiring the language model to generate a lemma specification to verify that it satisfies the specifications (e.g., that the assertion holds or does not result in an overflow). The language model generates proof and interacts with PISA. If proof is passed by Isabelle proving environment, we consider it a successful verification.

\noindent\textbf{Evaluation.}
We follow the evaluation settings of Lemur~\cite{wu2024lemur}. Within a specified timeout, Lemur, UAotumizer, and ESBMC generate proposals and call solvers for verification. Our approach interacts with PISA and self-corrects by the returned error messages.

\subsection{Compared Methods}
The methods we compare include the symbolic solvers: Uautomizer \cite{DBLP:conf/tacas/HeizmannCDEHLNSP13} and ESBMC \cite{DBLP:conf/kbse/GadelhaMMC0N18}, and the LLM-based method: Lemur \cite{wu2024lemur}.
UAUTOMIZER \cite{DBLP:conf/tacas/HeizmannCDEHLNSP13} is the overall champion of the 12th Competition on Software Verification (SV-COMP 2023). Combined with static analysis and model checking, it is one of the few verifiers that can give witness during verification.
ESBMC based on K-induction, which is particularly useful for verifying the properties of loops and recursive functions.
Lemur presents a set of derivation rules and makes proposals using a language model to approximate the boundary conditions of the loop invariant by interacting with the verifier.

\begin{table}[t]
  \caption{
      Result on formal verification task.
      FT: Fine-tuned.
  }
  \vspace{5pt}
  \label{tab:fv_result}
  \centering
  \begin{tabular}{lrrr}
    \toprule
    Model                                                               & Code2Inv (\#=133) & SV-COMP-47 (\#=47) & SV-COMP (\#=1,000) \\
    \midrule
    $\triangleright$ \textit{Symbolic Solver} \\
    \midrule
    UAUTOMIZER \cite{DBLP:conf/tacas/HeizmannCDEHLNSP13}                &  92   &  1   & 374 \\
    ESBMC \cite{DBLP:conf/kbse/GadelhaMMC0N18}                          &  68   &  1   & 358 \\    
    \midrule
    $\triangleright$ \textit{LLM-based Solver} \\
    \midrule
    Lemur-GPT-3.5-turbo \cite{wu2024lemur}                              &  103  &  14  & -  \\
    Lemur-GPT-4 \cite{wu2024lemur}                                      &  107  &  25  & - \\
    Mistral-7B \cite{DBLP:journals/corr/abs-2310-06825}                 &  37   &  10  & 75  \\
    Mistral-7B-FT                                                       &  40   &  14  & 84  \\
    Llama3-8B\footnotemark[4]                                           &  46   &  11  & 69  \\
    Llama3-8B-FT                                                        &  46   &  16  & 81  \\
    \bottomrule
  \end{tabular}
\end{table}

\begin{table}[t]
    \centering
    \caption{
        Failure types of Code2Inv and SV-COMP datasets.
    }
    \vspace{5pt}
    \begin{tabular}{lcccc}
    \toprule
    \multirow{2}{*}{Model} & \multicolumn{2}{c}{Code2Inv}          & \multicolumn{2}{c}{SV-COMP}   \\
                        \cmidrule(r){2-3} \cmidrule(r){4-5}
                              & statement error (\%) & proof error (\%) &  statement error (\%) & proof error (\%) \\
    \midrule
    Mistral-7B   & 70.8 & 29.2 & 49.7 & 50.3  \\
    Mistral-7B-FT & 72.0 & 28.0 & 59.0 & 41.0  \\
    Llama3-8B & 67.8 & 32.2 & 53.0 & 47.0  \\
    Llama3-8B-FT & 66.7 & 33.3 & 61.8 & 38.2  \\
    \bottomrule
    \end{tabular}
    \label{tab:fv_failure}
\end{table}

\subsection{Formal Verification Results}
Table~\ref{tab:fv_result} reports the number of passed verification tasks. Formal verification for C code in the Isabelle environment is a great challenge. First, the language model needs to generate the correct lemma specification, which is particularly difficult on code2inv and SV-comp-47 datasets with loops or complex conditions, and thus our fine-tuned prover model achieves limited performance gains. 
On the Code2Inv dataset, the uncertain looping conditions pose an additional challenge for the language model to validate C programs. 
The performance of the fine-tuned model on the SV-COMP-47 dataset equals or exceeds that of Lemur-GPT-3.5-trubo.
In addition, symbolic solvers overwhelmingly dominate the SV-comp-1,000 dataset, which covers diverse specifications. The lack of a relevant corpus makes it difficult for language models to verify specifications such as concurrency and no-overflow. Since \ourdataset\ originates from the seL4 micro-kernel operating system, the correlation of the data makes fine-tuning on the SV-COMP dataset effective.

\subsection{Ablation Study}
Further analysis in Table ~\ref{tab:fv_failure} shows that most of the validation errors in the Code2Inv dataset come from specification generation, which can be type mismatching, syntax errors, etc. In special, it is difficult to generate an accurate lemma specification under uncertain loop conditions. In contrast, the SV-COMP dataset has a larger fraction of validation errors from proof generation, and our finetuned prover model effectively reduces these proof errors. It suggests that it is feasible to utilize language models for formal verification in the Isabelle environment, but how to verify that the lemma specification generated by the model is semantically and syntactically correct remains a challenge.

\section{Conclusion}
This paper proposes \ourmethod, an interactive formal verification environment that can interact with LLMs by formulating formal verification (FV) dependencies and requests into automated theorem proving (ATP) theories and lemmas, and the verification processes into lemma proofs. 
We extract and cleanse a large-scale dataset \ourdataset\ with deep dependencies among Isabelle theorems and lemmas that formulate the formal verification.
Statistical analysis suggests that \ourdataset\ has comprehensive and deep dependency information among the theorems and lemmas, and the multi-step lemma proofs reach 100 steps. 
We benchmark \ourdataset\ by fine-tuning LLMs and then interacting with the \ourmethod\ environment. We evaluate Llama3-8B and Mistral-7B in this setting. 
Evaluations on Code2Inv and SV-COMP show improvements. For example, performances on SV-COMP of 17.39\% (69$\to$81) by Llama3-8B and 12\% (75$\to$84) by Mistral-7B, and the proof error proportions are reduced.
The results demonstrate the benefits of \ourmethod\ and \ourdataset.

\bibliography{FormalVer_references}
\bibliographystyle{plain}


\newpage
\appendix

\renewcommand{\contentsname}{Appendix}
\tableofcontents

\section{Limitations}
\label{sec:limitations}
In this work, we follow previous works \cite{DBLP:conf/tacas/HeizmannCDEHLNSP13,DBLP:conf/kbse/GadelhaMMC0N18,wu2024lemur} to test \ourmethod\ on C code verification. We remain the extension of \ourmethod\ and the corresponding \ourdataset\ to support more program languages as a near future work. Additionally, semantic alignment between lemma statements and program specifications is an unexplored area of research.

\section{Societal Impacts}
\label{sec:societal_impacts}
The research presented in this paper has the potential to advance the field of formal verification, automated theorem proving, AI for Math, and software engineering. The advancement can enhance the capabilities of large language models in formal verification, contributing to more reliable software development. By directly releasing the code and data,
we aim to ensure the responsible use of our work, fostering further innovation and maintaining high
standards of data privacy and intellectual property compliance. The proposed \ourmethod\ and \ourdataset\ benchmark the interactive formal verification performance in the machine learning field. Therefore we claim that there are no negative social impacts in this paper.

\section{\ourdataset\ Benchmark}

\subsection{Dataset Format}
We first list the folder and files under the \ourdataset\ directory.
We then demonstrate the detailed formats of the folder/files. 
\begin{itemize}
    \item \texttt{\hldbGray{sel4\_extraction/}} is a folder that has the same structure as the sel4 verification project (l4v). Each file is the extracted step-wise proof state of the corresponding l4v theory files. For example, ``\texttt{sel4\_extraction/proof/invariant-abstract/AInvs.json}'' is the proof state of the file \texttt{l4v/proof/invariant-abstract/AInvs.thy}.
    \item \texttt{\hldbGray{dataset\_lemma\_split.json}} contains all lemmas proof steps and states, and splits them into the train, val, test, and test-hard set.
    \item \texttt{\hldbGray{sel4\_thy\_info.json}} contains information of all theory files, including their names, dependency relations, and lemmas.
    \item \texttt{\hldbGray{sel4\_session\_info.json}} contains all session information, including dependent sessions, theories, and directories.
\end{itemize}

\subsubsection{\texttt{sel4\_extraction/}}
\noindent{The \texttt{sel4\_extraction/} folder} contains parsed l4v theory files. Each theory file in this folder is a JSON file, storing a list of whole proof steps, and each step is stored as a dictionary. 
The file structure and a sample proof step are demonstrated as follows:
\small
\begin{lstlisting}[language=json,numbers=none]
sel4_extraction/proof/invariant-abstract/AInvs.json: 
[
  ...,
  {        
    "index": 2,
    "step": "lemma st_tcb_at_nostate_upd: ...",
    "raw_output": "proof (prove)\ngoal (1 subgoal)...",
    "step_time": 0.11420297622680664
  },
  ...
]
\end{lstlisting}
\normalsize
Each proof step dictionary has the following fields:
\begin{itemize}
    \item ``\texttt{index}'': The index of this step.
    \item ``\texttt{step}'': The proof step in Isabelle. 
    \item ``\texttt{raw\_output}'': The returned proof state in Isabelle.
    \item ``\texttt{step\_time}'': The processing time of this step.
\end{itemize}

\subsubsection{\texttt{dataset\_lemma\_split.json}}
\noindent{The \texttt{dataset\_lemma\_split.json} file} stores the train/val/test/test-hard splits. Each split is a list of lemmas, and each is stored as a dictionary.
The file structure and a sample lemma are demonstrated as follows:
\small
\begin{lstlisting}[language=json,numbers=none]
{
  "train": [
    {
      "context": "lemma n_less_equal_power_2:\n  \"n < 2 ^ n\" by (fact less_exp)",
      "proof": [
        "lemma n_less_equal_power_2:\n  \"n < 2 ^ n\"",
        "by (fact less_exp)"
      ],
      "proof_state": [
        "proof (prove)\ngoal (1 subgoal):\n 1. n < 2 ^ n",
        ""
      ],
      "statement": "lemma n_less_equal_power_2:\n  \"n < 2 ^ n\"",
      "theory_name": "More_Arithmetic",
      "num_steps": 1
    },
    ...
  ],
  "val": [ ... ],
  "test": [ ... ],
  "test-hard": [ ... ] 
} 
\end{lstlisting}
\normalsize
Each lemma dictionary has the following fields:
\begin{itemize}
    \item ``\texttt{context}'': Full lemma context in plain text.
    \item ``\texttt{proof}'': A list of all proof steps in Isabelle.
    \item ``\texttt{proof\_state}'': A list of all proof states in Isabelle.
    \item ``\texttt{statement}'': The lemma statement to be proved.
    \item ``\texttt{theory\_name}'': The name of the theory where this lemma belongs.
    \item ``\texttt{num\_steps}'': The number of steps for proving this lemma.
\end{itemize}

\subsubsection{\texttt{sel4\_thy\_info.json}}
\noindent\textbf{\texttt{sel4\_thy\_info.json}} contains information regarding the theory files, stored as a dictionary where a key is a theory file and the value contains the related information. A sample is demonstrated as follows:
\small
\begin{lstlisting}[language=json,numbers=none]
{
  ...,
  "/lib/Word_Lib/More_Word.thy": {
    "name": "More_Word",
    "dependency": {
      "HOL-Library.Word": "",
      "More_Arithmetic": "/lib/Word_Lib",
      "More_Divides": "/lib/Word_Lib",
      "More_Bit_Ring": "/lib/Word_Lib"
    },
    "depth": 2,
    "related_c_code": [],
    "child": [
      "/lib/Word_Lib/Aligned.thy",
      "/lib/Word_Lib/Bit_Shifts_Infix_Syntax.thy",
      ...,
      "/lib/Word_Lib/Machine_Word_64.thy"
    ],
    "path": "/lib/Word_Lib/More_Word.thy",
    "session": "Word_Lib",
    "lemmas": [
      {
        "context": "lemma sofl_test: ...",
        "proof": [...],
        "proof_state": [...],
        "statement": "...",
        "theory_name": "More_Word",
        "num_steps": 25
      },
  },
  ...
}
\end{lstlisting}
\normalsize
The information dictionary of a theory file (e.g., ``\texttt{/lib/Word\_Lib/More\_Word.thy}'') has the following fields:
\begin{itemize}
    \item ``\texttt{name}'': The theory name.
    \item ``\texttt{dependency}'': A dictionary of dependent theories and their paths. The key is the theory name and the value is the path. A theory that belongs to another session has no path. For example, ``\texttt{HOL-Library.Word}'' is imported from session ``\texttt{HOL-Library}'', and its path is empty.
    \item ``\texttt{depth}'': The depth of this theory.
    \item ``\texttt{related\_c\_code}'': The C code files called by this theory or any of its ancestors. 
    \item ``\texttt{child}'': The theory files depending on this theory.
    \item ``\texttt{path}'': The theory file path relative to the l4v folder.
    \item ``\texttt{session}'': The session that contains this theory.
    \item ``\texttt{lemmas}'': The list of all lemmas in this theory files. Each lemma is stored in a dictionary, which is the same as in ``\texttt{dataset\_lemma\_split.json}''. 
\end{itemize}

\subsubsection{\texttt{sel4\_session\_info.json}}
\noindent\textbf{\texttt{sel4\_session\_info.json}} contains information regarding each l4v session, stored as a dictionary where a key is an l4v session and the value contains the related information. 
A sample is demonstrated as follows:
\small
\begin{lstlisting}[language=json,numbers=none]
{
  "ASpec": {
    "dependency": [
      "Word_Lib",
      "\"HOL-Library\"",
      "Lib",
      "ExecSpec"
    ],
    "name": "ASpec",
    "theories": [
      "/spec/abstract/Structures_A.thy",
      ...,
      "/spec/abstract/Exceptions_A.thy"
    ],
    "ROOT_dir": "/spec",
    "ROOT_relative_dir": "abstract",
    "additional_dir": [
      ".",
      "ARM"
    ],
    "depth": 6
  },
  ...
}
\end{lstlisting}
\normalsize
The information dictionary of a session (e.g., ``\texttt{ASpec}'') has the following fields:
\begin{itemize}
    \item ``\texttt{dependency}'': A list of all its dependent sessions' names.

    \item ``\texttt{name}'': The session name.

    \item ``\texttt{theories}'': The list of all theory files included in this session, represented by their keys in ``\texttt{sel4\_thy\_info.json}''.

    \item ``\texttt{ROOT\_dir}'': The directory of this session's ROOT file relative to the l4v folder.
    
    \item ``\texttt{ROOT\_relative\_dir}'': The main working directory of this session relative to ``\texttt{ROOT\_dir}''.

    \item ``\texttt{additional\_dir}'': The list of additional directories containing this session's theory files relative to ``\texttt{ROOT\_relative\_dir}''.

    \item ``\texttt{depth}'': The depth of this session. 
\end{itemize}

\subsection{Datasheet}
We present a datasheet \cite{DBLP:journals/cacm/GebruMVVWDC21} for documentation and responsible usage of \ourdataset\ benchmark.

\paragraph{Motivation.}
\begin{itemize}
\item \textit{For what purpose was the dataset created?} 
The \ourdataset\ dataset is created to support the interactive formal verification with large language models. It provides lemmas for formally proofing the correctness of a microkernel system with step-wise Isabelle language and state.
\item \textit{Who created the dataset (e.g., which team, research group) and on behalf of which entity (e.g., company, institution, organization)?} 
It was created by the authors of this paper by extracting and cleansing the data from the \hyperlink{https://github.com/seL4/l4v}{sel4 verification project (l4v)}.
\item \textit{Who funded the creation of the dataset?} 
See the acknowledgments once it is available.
\end{itemize}

\paragraph{Composition.}
\begin{itemize}
\item \textit{What do the instances that comprise the dataset represent (e.g., documents, photos, people, countries)?} 
The \ourdataset\ dataset consists of dependent theory sessions, theory files grouped by sessions, lemmas from theories, and proof states of the lemmas, all written in Isabelle.
\item \textit{How many instances are there in total (of each type, if appropriate)?} The \ourdataset\ dataset has 758 theories, 29,125 lemmas, and 200,646 proof steps. 
\item \textit{Does the dataset contain all possible instances or is it a sample (not necessarily random) of instances from a larger set?} The dataset contains all possible theory files, lemma, and their proof that PISA can extract from the \hyperlink{https://github.com/seL4/l4v}{sel4 verification project (l4v)} in ARM architecture(excluding C Parser and autocorres tools) released on March 11, 2024.
\item \textit{What data does each instance consist of?} Each instance consists of the lemma statement, the proof step, and the corresponding state in Isabelle code.
\item \textit{Is there a label or target associated with each instance?} Yes, each instance has a target, the next proof step.
\item \textit{Is any information missing from individual instances?} No.
\item \textit{Are relationships between individual instances made explicit (e.g., users' movie ratings, social network links)?} Yes, each instance is associated with a theory file, which contains dependent theory files as its premises.
\item \textit{Are there recommended data splits (e.g., training, development/validation, testing)?} Yes. We recommend four data splits: a training set with 26,081 lemmas, a validation set with 1,115 lemmas, a test set with 1,077 lemmas, and a test-hard set with 852 lemmas. 
\item \textit{Are there any errors, sources of noise, or redundancies in the dataset?} The extracted lemma is formally verified by Isabelle and thus has no error or noise. There might exist some redundant proof that is very similar to the others.
\item \textit{Is the dataset self-contained, or does it link to or otherwise rely on external resources (e.g., websites, tweets, other datasets)?} The dataset is self-contained. 
\item \textit{Does the dataset contain data that might be considered confidential (e.g., data that is protected by legal privilege or by doctor-patient confidentiality, data that includes the content of individuals’ non-public communications)?} No.
\item \textit{Does the dataset contain data that, if viewed directly, might be offensive, insulting, threatening, or might otherwise cause anxiety?} No. 
\end{itemize}

\paragraph{Collection Process.}
\begin{itemize}
    \item \textit{How was the data associated with each instance acquired?} The original data contains Isabelle theory files structured with ROOT file. We apply \ourmethod to extract their proof steps and states. The details are described in Section 4 of our paper.
    \item \textit{What mechanisms or procedures were used to collect the data (e.g., hardware apparatuses or sensors, manual human curation, software programs, software APIs)?} The original data is publicly released in \hyperlink{https://github.com/seL4/l4v}{https://github.com/seL4/l4v}.
    \item \textit{Who was involved in the data collection process (e.g., students, crowdworkers, contractors) and how were they compensated (e.g., how much were crowdworkers paid)?} No manual effort was involved in the data collection process.
    \item \textit{Over what timeframe was the data collected?} The dataset was collected on March 11, 2024. 
\end{itemize}

\paragraph{Preprocessing/cleaning/labeling.}
\begin{itemize}
    \item \textit{Was any preprocessing/cleaning/labeling of the data done (e.g., discretization or bucketing, tokenization, part-of-speech tagging, SIFT feature extraction, removal of instances, processing of missing values)?} The original l4v theory file is parsed into step-wise language by Isabelle. We then interact with Isabelle using these steps to obtain the step-wise states.
    \item \textit{Was the ``raw'' data saved in addition to the preprocessed/cleaned/labeled data (e.g., to support unanticipated future uses)?} Yes. We store the original seL4 formal verification files used for extraction and record the links between each lemma and its original files.
    \item \textit{Is the software that was used to preprocess/clean/label the data available?} Yes. We release the codes and environments for extracting seL4 formal proofs.
\end{itemize}

\paragraph{Uses.}
\begin{itemize}
    \item \textit{Has the dataset been used for any tasks already?} We have used the dataset for fine-tuning Mistral-7B and llama3-8B for the \ourmethod\ environment. We also use the dataset to evaluate the fine-tuned models.
    \item \textit{Is there a repository that links to any or all papers or systems that use the dataset?} \url{https://fveler.github.io/}.
    \item \textit{What (other) tasks could the dataset be used for?} The dataset can be used for pertaining LLMs for various downstream tasks, such as ATP, MWP, and code generation.
    \item \textit{Is there anything about the composition of the dataset or the way it was collected and preprocessed/cleaned/labeled that might impact future uses?} The dataset is based on l4v and is extracted with Isabelle 2023. The lemma proof and proof states might be different from future versions of l4v or incompatible with future versions of Isabelle.
    \item \textit{Are there tasks for which the dataset should not be used?} No. 
\end{itemize}

\paragraph{Distribution.}
\begin{itemize}
    \item \textit{Will the dataset be distributed to third parties outside of the entity (e.g., company, institution, organization) on behalf of which the dataset was created?} Yes, the dataset is publicly available on the Internet. 
    \item \textit{How will the dataset will be distributed (e.g., tarball on website, API, GitHub)?} The dataset can be downloaded as a tarball.
    \item \textit{When will the dataset be distributed?} 
    The dataset has been released and can be downloaded from \url{https://huggingface.co/FVELer}.
    \item \textit{Will the dataset be distributed under a copyright or other intellectual property (IP) license, and/or under applicable terms of use (ToU)?} The dataset is distributed under CC BY 2.0. The dataset was extracted from the \hyperlink{sel4 verification project (l4v)}{https://github.com/seL4/l4v} and is licensed under GPL version 2. 
    \item \textit{Have any third parties imposed IP-based or other restrictions on the data associated with the instances?} No.
    \item \textit{ Do any export controls or other regulatory restrictions apply to the dataset or to individual instances?} No.
\end{itemize}

\paragraph{Maintenance.}
\begin{itemize}
    \item \textit{Who will be supporting/hosting/maintaining the dataset?} The authors of this paper. 
    \item \textit{How can the owner/curator/manager of the dataset be contacted
(e.g., email address)?} Please
contact Qingxing Cao at caoqx8@sysu.edu.cn.
    \item \textit{Is there an erratum?} No.
    \item \textit{Will the dataset be updated (e.g., to correct labeling errors, add
new instances, delete instances)?} Please check \url{https://https://fveler.github.io/} for any update. 
    \item \textit{If others want to extend/augment/build on/contribute to the dataset, is there a mechanism for them to do so?} 
    Yes. they can use our released data extraction code for extending instances from updated seL4 or other related data sources.
\end{itemize}

\textbf{\subsection{Data Hosting, Licensing, and Maintenance}}
\ourdataset\ benchmark is distributed under the CC BY 2.0 license. 
The data and the corresponding documentation are hosted on Hugging Face at \url{https://huggingface.co/FVELer}. 
The codes for data extraction and experiments with the corresponding documentation are released at \url{https://github.com/FVELER/FVEL}. 
The model checkpoints are hosted on Hugging Face Hub.
Our website of \ourmethod\ and \ourdataset\ is \url{https://fveler.github.io/}, which is the entry point for everything related, including future updates or maintenance.

\begin{table*}[!t]
    \scriptsize
    \caption{
    Demonstrations of input-target pairs in alpaca format for fine-tuning \ourmethod-LLMs. 
    }
    \label{tab:example}
    \begin{tabular}{
        >{\color{black}}p{\textwidth}
    }
        \toprule \midrule
        \vspace{-4pt}\textbf{\textsf{Instruction}} \\
        \vspace{-4pt}\texttt{Prove the following lemma statement in Isabelle.\newline Ensure that the proof is complete, logically sound and free of redundant content. Use appropriate tactics and lemmas as necessary. Don't explain.}\\
        \vspace{-4pt}\textbf{\textsf{Input}} \\
        \vspace{-4pt}\texttt{lemma pred\_restrs\_cut:
  "($\exists$y. restrs x $\subseteq$ {y}) $\Rightarrow$ ($\exists$y. pred\_restrs nn restrs x $\subseteq$ {y})"} \\   
        \vspace{-4pt}\textbf{\textsf{output}} \\   
        \vspace{-4pt}\texttt{apply (clarsimp simp: pred\_restrs\_def split: next\_node.split) apply blast done} \\
        \midrule \midrule 
        \vspace{-4pt}\textbf{\textsf{Instruction}} \\
        \vspace{-4pt}\texttt{Prove the following lemma statement in Isabelle.\newline Ensure that the proof is complete, logically sound and free of redundant content. Use appropriate tactics and lemmas as necessary. Don't explain.}\\
        \vspace{-4pt}\textbf{\textsf{Input}} \\
        \vspace{-4pt}\texttt{lemma if\_x\_None\_eq\_Some:
  "((if P then x else None) = Some y) = (P $\wedge$ x = Some y)"} \\
        \vspace{-4pt}\textbf{\textsf{Output}} \\ 
        \vspace{-4pt}\texttt{by simp} \\
        \midrule \midrule 
        \vspace{-4pt}\textbf{\textsf{Instruction}} \\
        \vspace{-4pt}\texttt{Prove the following lemma statement in Isabelle.\newline Ensure that the proof is complete, logically sound and free of redundant content. Use appropriate tactics and lemmas as necessary. Don't explain.}\\
        \vspace{-4pt}\textbf{\textsf{Input}} \\
        \vspace{-4pt}\texttt{lemma visit\_merge: \newline
  assumes tr: "tr $\in$ exec\_trace Gamma fn" "Gamma fn = Some f"\newline
      and wf: "wf\_graph\_function f ilen olen"\newline
      and ns: "nn $\neq$ NextNode (entry\_point f)"\newline
        "$\forall$n $\in$ set ns. graph n = Some (Basic nn [])"\newline
        "converse (reachable\_step graph) `` \{nn\} $\subseteq$ NextNode ` set ns"\newline
     and geq: "function\_graph f = graph"\newline
     and cut: "$\forall$x. NextNode x $\in$ set cuts $\rightarrow$ ($\exists$y. restrs x $\subseteq$ \{y\})"\newline
              "$\forall$n $\in$ set ns. (nn, NextNode n) $notin$ rtrancl\newline
                      (reachable\_step graph $\bigcap$ \{(x, y). x $notin$ set cuts\})"\newline
  shows "visit tr nn restrs = fold merge\_opt (map ($\lambda$n. visit tr (NextNode n)\newline
            (pred\_restrs' n restrs)) ns) None"} \\
        \vspace{-4pt}\textbf{\textsf{Output}} \\ 
        \vspace{-4pt}\texttt{proof - note ns = ns[folded geq] note cut = cut[folded geq] have step\_after:\newline
    "$\wedge$n i. n $\in$ set ns $\Rightarrow$ trace\_bottom\_addr tr i = Some (NextNode n)\newline
        $\Rightarrow$ $\exists$st. tr i = Some [(NextNode n, st, fn)]\newline
            $\wedge$ tr (Suc i) = Some [(nn, st, fn)]\newline
            $\wedge$ trace\_addr tr (Suc i) = Some nn\newline
            $\wedge$ restrs\_condition tr restrs (Suc i)\newline
                = restrs\_condition tr (pred\_restrs' n restrs) i" apply (drule exec\_trace\_non\_Call[OF tr], (simp add: ns)+) apply (frule ns[rule\_format], cut\_tac tr(2)) apply (frule trace\_addr\_SomeD, clarsimp) apply (frule exec\_trace\_invariant[OF tr(1)]) apply (cut\_tac i=i in exec\_trace\_step\_cases[OF tr(1)]) apply (clarsimp simp: all\_exec\_graph\_step\_cases exec\_graph\_invariant\_Cons \newline
                          upd\_vars\_def save\_vals\_def) apply (simp add: pred\_restrs[OF tr(1)] trace\_addr\_SomeI trace\_bottom\_addr\_def K\_def) done have step\_after\_single: \newline
    "$\wedge$n i. n $\in$ set ns $\Rightarrow$ trace\_bottom\_addr tr i = Some (NextNode n) \newline
        $\Rightarrow$ restrs\_condition tr restrs (Suc i) \newline
        $\Rightarrow$ ($\forall$n' j. n' $\in$ set ns $\rightarrow$ trace\_addr tr j = Some (NextNode n') \newline
                $\rightarrow$ restrs\_condition tr (pred\_restrs' n' restrs) j $\rightarrow$ j = i)" apply clarsimp apply (frule step\_after, erule trace\_addr\_trace\_bottom\_addr\_eq) apply (frule(1) step\_after) apply clarsimp apply (drule(2) restrs\_single\_visit[OF tr wf \_ \_ \_ \_ cut(1)], simp\_all) apply (rule not\_trancl\_converse\_step, rule ns) apply (simp add: cut) done have visit\_after: \newline
    "$\wedge$n v. n $\in$ set ns $\Rightarrow$ visit tr (NextNode n) (pred\_restrs' n restrs) = Some v \newline
        $\Rightarrow$ visit tr nn restrs $\neq$ None" apply (clarsimp simp: visit\_eqs) apply (drule\_tac i=i in step\_after, simp add: trace\_addr\_trace\_bottom\_addr\_eq) apply (rule\_tac x="Suc i" in exI) apply clarsimp done show ?thesis apply (rule sym, cases "visit tr nn restrs", simp\_all) apply (rule fold\_merge\_opt\_Nones\_eq) apply (rule ccontr, clarsimp simp: visit\_after) apply (clarsimp simp: visit\_eqs) apply (frule visit\_immediate\_pred[OF tr wf \_ ns(1, 3)]) apply clarsimp apply (frule(1) step\_after, clarsimp) apply (frule(2) step\_after\_single) apply (drule in\_set\_conv\_decomp\_last[THEN iffD1]) apply clarsimp apply (rule trans, rule fold\_merge\_opt\_Nones\_eq) apply (rule ccontr, clarsimp simp: visit\_eqs pc\_def ball\_Un) apply (simp add: trace\_addr\_SomeI) apply (subst visit\_known, assumption, simp\_all) apply clarsimp done qed}
        \\
        \midrule \bottomrule
    \end{tabular}    
\end{table*}

\section{Experiments on \ourdataset\ Test Set}
\subsection{Implementation Details}
\label{app: details}
We use Mistral-7B-Instruct-v0.2\footnote{\url{https://huggingface.co/mistralai/Mistral-7B-Instruct-v0.2}} \cite{DBLP:journals/corr/abs-2310-06825} and LLama-3-8B-Instruct models\footnote{\url{https://github.com/meta-llama/llama3}} to conduct the experiments on \ourdataset\ test sets. 

\paragraph{Fine-tuning.} 
We use the LLaMa-Factory \cite{zheng2024llamafactory} framework to fine-tune two models on a single RTX 3090 GPU. Specifically, we deploy LORA\cite{hu2021lora} on the q\_proj and v\_proj modules of both models. We filter out training samples of a length greater than 1024 and feed the remaining samples into the model with a global batch size of 8. The training samples are transformed into the alpaca format, demonstrated in Table~\ref{tab:example}.

We use a cosine learning rate scheduler with the maximum $lr=1e-4$, minimal $lr=1e-5$, $warmup\_ratio=0.1$. Fine-tuning lasts up to 10 epochs with an early-stop mechanism at minimal evaluation loss. For Mistral-7B and Llama-3-8B, the fine-tuning lasts about 30 GPU hours.

\paragraph{Inference.} 
During inference, the model generates the complete proof in a single pass using a zero-shot approach, and PISA verifies the resulting proof. 
We manually add the imports and include theorems needed for the proof to the environment so that they are correctly referenced. 
We set the temperature to 0.0 during inference to assess the model's greedy performance.
The inference template to prompt the model is demonstrated in Table~\ref{tab:inference_prompt}.
\begin{table}[t]
    \centering
    \caption{Inference template for prompting \ourmethod-LLMs.}
    \label{tab:inference_prompt}
    \begin{tabular}{
    >{\raggedright\arraybackslash}P{.97\textwidth}
    }
    \toprule \midrule
        \texttt{Prove the following lemma statement in Isabelle.
        Ensure that the proof is complete, logically sound and free of redundant content. Use appropriate tactics and lemmas as necessary. Don't explain.} \\

        \texttt{\{statement\}} \\
    \midrule \bottomrule
    \end{tabular}    
\end{table}


\subsection{Results}
Table~\ref{tab:atp_result} illustrates the result of fine-tuning Mistral and Llama3 on our \ourdataset\ training set and testing on the \ourdataset\ test set and test-hard set. The fine-tuned Llama-3-8B and mistral-7B effectively improve the correctness of the proofs, with FVEL-Mistral-7B and FVEL-Llama-3-8B each achieving a 4.5\% improvement (2.4\% -> 6.9\% and 3.6\% -> 8.1\%, respectively) on the \ourdataset\ test split. On the more complex \ourdataset\ test-hard split, 3.5\% (2.3\% -> 5.8\%) and 4.3\% (3.2\% -> 7.5\%) improvement are achieved respectively. 
Currently, the pass rate for both Mistral and Llama remains relatively low, indicating that the proposed benchmark poses significant challenges for LLMs.
The poor results are primarily caused by these two factors:
1) \textbf{Data scarcity.} The amount of data available on formal verification is relatively small compared to the data required to train a general LLM. This is a long-standing challenge in the domain of formal mathematics and formal verification. \ourdataset\ remedies the issue by incorporating data from formal verification, but we still require much more data for the LLM to perform better on the subject.
2) \textbf{Tactic application style.} The majority of proofs are written in a tactic application style. Compared to the declarative style, these codes cannot be understood even by humans without interacting with Isabelle and checking the proof state information given by the formal system. The current whole proof paradigm requires generating the proof in one go without the help of the proof state information, which poses a significant challenge.

\begin{table}[t]
    \centering
    \caption{Automated theorem proving result on \ourdataset.}
    \begin{tabular}{lcc}
    \toprule
                           & \ourdataset\ test (\#=1077) & \ourdataset\ test-hard (\#=852) \\
    \midrule
     Mistral-7B            & 26                         & 19 \\
     \ourmethod-Mistral-7B & 74                         & 49 \\
     Llama-3-8B            & 39                         & 27 \\
     \ourmethod-Llama-3-8B & 88                         & 64 \\
     \bottomrule
    \end{tabular}
    
    \label{tab:atp_result}
\end{table}

\begin{table*}[!t]
    \scriptsize
    \caption{
    Compared generated proofs by Mistral-7B and \ourmethod-Mistral-7B.
    }
    \label{tab:atp_cases}
    \begin{tabular}{
        P{.2\textwidth}
        >{\raggedright\arraybackslash}P{.35\textwidth}
        >{\raggedright\arraybackslash}P{.35\textwidth}
    }
        \toprule
        \textbf{\textsf{Input}} & \textbf{\textsf{Mistral-7B}} & \textbf{\textsf{\ourmethod-Mistral-7B}} \\
        \midrule
        lemma not\_minus: \newline  "NOT (x - y) = y - x - 1"
        & 
        assume "NOT (x - y) = y - x - 1" \newline  hence "x - y = y - x - 1" by simp \newline  hence "x = y - 1" by simp \newline  
        thus "x = y" by contradiction \newline 
        qed
        &
        by (simp add: not\_eq\_complement) \\
        \midrule
        lemma and\_and\_not [simp]: "(a AND b) AND NOT b = 0" for a b :: "'a::len word"
        & 
        lemma and\_and\_not[simp]: \newline
          assumes "a: 'a::len word" and "b: 'a::len word"  \newline 
          shows "(a AND b) AND NOT b = 0" \newline
        proof - \newline
          from assms have "a AND b = (a AND b) AND (a OR b)" by simp \newline
          then have "a AND (b AND NOT b) = (a AND b) AND (a OR b)" by simp \newline
          finally show ?thesis by simp \newline
        qed
        & 
        proof - \newline
          have "(a AND b) AND NOT b = a AND (b AND NOT b)" \newline
            by (simp add: word\_bw\_assoc) \newline
          also have "... = a AND 0" \newline
            by (simp add: word\_and\_not) \newline
          also have "... = 0" \newline
            by simp \newline
          finally show ? thesis \newline
            by simp \newline
        qed 
        \\
        \bottomrule
    \end{tabular}    
\end{table*}

\subsection{Case Study}
Table~\ref{tab:atp_cases} demonstrates compared generated proofs by Mistral-7B and \ourmethod-Mistral-7B after being fine-tuned with \ourdataset. 
%
The upper row shows a case in which FVEL-Mistral-7B correctly applies the lemma learned from fine-tuning, thus correcting and simplifying the proof. Contrastively, Mistral-7B generates common \texttt{not\_eq\_complement} without considering a reasonable proof strategy, resulting in a failed proof.
%
In the second case, Mistral-7B rewrites the lemma statement into ``\texttt{assumes}'' and ``\texttt{shows}'' statements, according to which gives an incorrect proof.
FVEL-Mistral-7B, on the other hand, expands the brackets in the equation and then is able to derive contradiction according to ``(b AND NOT b)'', and completes the proof via the contradiction of the right-hand side of the equation.

\begin{table}[t!]
  \scriptsize
  \centering
  \caption{Comparison of Original and Processed C Code}
  \label{tab:compare_code}
  \begin{tabular}{|p{0.45\textwidth}|p{0.45\textwidth}|}
    \hline
    \multicolumn{1}{|c|}{\textbf{Original Code}} & \multicolumn{1}{c|}{\textbf{Processed Code}} \\
    \hline
    \begin{lstlisting}[language=c]
extern void abort(void);
extern void __assert_fail(const char *, const char *, unsigned int, const char *) __attribute__ ((__nothrow__ , __leaf__)) __attribute__ ((__noreturn__));
void reach_error() { __assert_fail("0", "nested3-2.c", 3, "reach_error"); }

void __VERIFIER_assert(int cond) {
  if (!(cond)) {
    ERROR: {reach_error();abort();}
  }
  return;
}

int main()
{
  unsigned int x = 0;
  unsigned int y = 0;
  unsigned int z = 0;
  unsigned int w = 0;

  while (x < 0x0fffffff) {
    y = 0;

    while (y < 0x0fffffff) {
    z =0;
  while (z <0x0fffffff) {
    z++;
  }
    __VERIFIER_assert(!(z % 4));
  y++;
    }
    __VERIFIER_assert(!(y % 2));

    x++;
  }
  __VERIFIER_assert(!(x % 2));
 return 0;

}
    \end{lstlisting} &
    \begin{lstlisting}[language=c]
extern void abort(void);

void VERIFIER_assert(int cond) {
  if (!(cond)) {
    {abort();}
  }
  return;
}

int main()
{
  unsigned int x = 0;
  unsigned int y = 0;
  unsigned int z = 0;
  unsigned int w = 0;

  while (x < 0x0fffffff) {
    y = 0;

    while (y < 0x0fffffff) {
    z =0;
  while (z <0x0fffffff) {
    z++;
  }
    VERIFIER_assert(!(z % 4));
  y++;
    }
    VERIFIER_assert(!(y % 2));

    x++;
  }
  VERIFIER_assert(!(x % 2));
 return 0;

}
    \end{lstlisting} \\
    \hline
  \end{tabular}
\end{table}

\section{Implementations Details on Code2Inv and SV-COMP}
This section provides supplementary details regarding the benchmark study in Section 5.

\subsection{Evaluation Datasets}
\noindent\textbf{Code2Inv~\cite{DBLP:conf/cav/SiNDNS20}.} The code2inv dataset contains 133 programs in c, each containing a pre-condition, a loop body (while or for statement), and a post-condition. The verifier needs to verify that the post-condition (an assertion) holds. It is worth pointing out that the condition of a loop or branch in the program may be indeterminate.

\noindent\textbf{SV-COMP~\cite{DBLP:conf/tacas/Beyer23}.} The Software-Verification Competition provides a diverse set of benchmarks for formal verification. sv-comp benchmark contains over 23k c programs, which tend to be more complex than those in code2inv, and each program is accompanied by a .yml file to declare its specifications. These specifications cover requirements such as ReachSafety, MemSafety, ConcurrencySafety, NoOverflows, Termination, etc. The verifier is required to determine whether a program satisfies the given specifications. We sampled the SV-COMP benchmark into two subsets: a 47-sample subset sampled by Lemur \cite{wu2024lemur}, which contains samples with multiple nested loops, and a 1,000-sample subset which is randomly sampled from the full set. In particular, we exclude samples that contain floating-point type because the C-parser cannot parse them correctly.

\subsection{Pre-processing}
Table~\ref{tab:compare_code} demonstrates a randomly selected sample before pre-processing (original code) and after pre-processing (processed code).
The pre-processing stages are explained as follows.

\paragraph{Data Preprocess.}
Since C-parser supports only part of the C99 standard, some C features (e.g. ``\texttt{goto}'' statements, side effects in expressions, etc.) are not supported, we normalize the C code to make C-parser work properly. Specially, for C code which includes:

\noindent\textbf{String Literal and Illegal Function Name.} Functions with string literals are often used to give warnings to the verifier, we remove these functions and keep only ``\texttt{extern void abort(void);}'' 
In addition, we fix illegal function names, for example, by removing the underlines at the beginning of the name.

\noindent\textbf{Assertion and Assumption.} We replace all the ``\texttt{assert(statement);}'' and ``\texttt{assume(statement);}'' with ``\texttt{if (not (statement) \{return -1;\}}''. Note that all assertions appear in the ``\texttt{main()}'' function, so the semantics before and after the replacement are equivalent.

\noindent\textbf{Unknown Condition.} ``\texttt{unknown()}'' is often used in the Code2Inv dataset as a condition in ``\texttt{while}'' or ``\texttt{if}'' expressions, and we add external declarations to this function: ``\texttt{extern int unknown(void);}''.

\subsection{Fine-tuning and Inference} 
See Appendix~\ref{app: details} for fine-tuning and inference details.

\end{document}